\pdfoutput=1
\documentclass{article}

\usepackage{microtype}
\usepackage{graphicx}
\usepackage{subfigure}
\usepackage{booktabs} 
\usepackage{amsfonts,times,natbib}

\usepackage{bm}
\usepackage{amsmath, amssymb, theorem}
\theorembodyfont{\upshape}
\newtheorem{theo}{Theorem}
\newtheorem{defi}{Definition}
\newtheorem{prob}{Problem}
\newtheorem{lemm}{Lemma}
\newtheorem{prop}{Proposition}

\newtheorem{assu}{Assumption}
\def\qed{\hfill {$\Box$}}
\newenvironment{prf}{\par
 \vspace{0.1\topsep}\noindent{{\it Proof.} \rule{.15em}{0mm}}
}{
 \hfill{\qed}\par\vspace{0.8\topsep}
}

\newcommand{\supp}{\mathop{\textrm{supp}}}
\newcommand{\svm}{{\cal A}_{\textrm{SVM}}}
\newcommand{\sol}{{\textrm{SVM}}}
\newcommand{\set}[1]{\{#1\}}
\newcommand{\alp}{\bm{\alpha}}
\newcommand{\halp}{\hat \alp}

\newcommand{\name}[1]{{\em #1\/}}

\usepackage{hyperref}

\usepackage[accepted]{hill2019}

\icmltitlerunning{Enumeration of Distinct Support Vectors}

\begin{document}

\twocolumn[
\icmltitle{Enumeration of Distinct Support Vectors for Interactive Decision Making
            }



\icmlsetsymbol{equal}{*}

\begin{icmlauthorlist}
  \icmlauthor{Kentaro Kanamori}{hu}
  \icmlauthor{Satoshi Hara}{ou}
  \icmlauthor{Masakazu Ishihata}{ntt}
  \icmlauthor{Hiroki Arimura}{hu}
\end{icmlauthorlist}

\icmlaffiliation{hu}{Hokkaido University, Japan}
\icmlaffiliation{ou}{Osaka University, Japan}
\icmlaffiliation{ntt}{NTT Communication Science Laboratories, Japan}

\icmlcorrespondingauthor{Kentaro Kanamori}{kanamori@ist.hokudai.ac.jp}

\icmlkeywords{Model enumeration, Support Vector Machines (SVM), Example-based explanation}

\vskip 0.3in
]



\printAffiliationsAndNotice{}  

\begin{abstract}
  In conventional prediction tasks, a machine learning algorithm outputs a single best model that globally optimizes its objective function, which typically is accuracy.
  Therefore, users cannot access the other models explicitly.
  In contrast to this,
  multiple model enumeration attracts increasing interests in non-standard machine learning applications where other criteria, e.g., interpretability or fairness, than accuracy are main concern and a user may want to access more than one non-optimal, but suitable models.
  In this paper,
  we propose a $K$-best model enumeration algorithm for
  Support Vector Machines (SVM) that
  given a dataset $S$ and an integer $K>0$,
  enumerates the $K$-best models on $S$ with distinct support vectors
  in the descending order of the objective function values in the dual SVM problem.
  Based on analysis of the lattice structure of support vectors, our algorithm efficiently finds the next best model with small latency. This is useful in supporting users's interactive examination of their requirements on enumerated models.
  By experiments on real datasets, we evaluated the efficiency and usefulness of our algorithm.

\end{abstract}

\section{Introduction}
Machine learning technologies are being widely applied to decision making in the
real
world.
Recently, non-standard learning problems with criteria, such as \name{interpretability}~\cite{DBLP:conf/kdd/Ribeiro0G16,Angelino:2017}
and \name{fairness}~\cite{Hajian:2016:ABD:2939672.2945386,nips:2017:bias}, other than prediction accuracy attract increasing attention.
In case that the predictions by a
learning algorithm are not suitable to user's requirements, or violate critical constraints, it may no longer be usable in the actual world, even if it has high prediction accuracy.

To incorporate user's requirements into learning process,
a new framework, called \textit{model enumeration}, is recently proposed~\cite{hara:maehara:aaai:2017,Ruggieri:ICML:2017,hara:ishihata:AAAI:2018}.
In this framework, an algorithm enumerates several models with different structures, possibly with the same objective values,
instead of finding a single, optimal model.
It has
a number of
advantages to enumerate models.
The previous work~\cite{hara:maehara:aaai:2017} studied model enumeration
focusing on enumeration of subsets of \name{features}.
In contrast to this, we focus on
enumeration of distinct models based on subsets of \name{examples} in a given dataset.

In this study, we propose an enumeration algorithm for Support Vector Machines (SVM)~\cite{Vapnik:1998}.
In the dual form of the SVM learning problem, its decision boundary, i.e., its model is represented by
a linear combination of the subset of a given dataset, which is called \textit{support vectors}.
Adopting
the dual form of the SVM learning problem
and extending the enumeration method
for Lasso by~\cite{hara:maehara:aaai:2017},
we present an algorithm for enumerating SVM models that have distinct support vectors
in the descending order of the dual form objective function values.
Our approach has the following advantages:

\vspace{-2.5mm}
\begin{itemize}
  \itemsep=0pt
  \item \textbf{Data understanding: }
    A single model that optimizes its objective function
    is not necessarily the best model that can explain the data well,
    due to, e.g., label noise or data contamination.
    By enumerating many models,
    we have a chance to access better models from the user's interests.
    This can be seen as a multiple version of \textit{example-based explanation}~\cite{CIS-260113}.

  \item \textbf{Interactive learning: }
    In a long-term prediction service, a single optimal model may not continue to be the best model forever
    due to change of a user's interests or requirements.
    Our framework can be used to provide
    the next best model by a user's request to interactively examine and select some of enumerated models.

\end{itemize}


\subsubsection*{Contributions}
In this paper, we make the following contributions.
\vspace{-2.5mm}
\begin{enumerate}
  \itemsep=0pt
  \item We formulate a model enumeration problem for SVM
    as enumeration of SVM models with distinct support vectors
    in the descending order of the objective values.


  \item We propose an efficient exact algorithm for the SVM model enumeration problem
    by extending the approach for Lasso by~\cite{hara:maehara:aaai:2017}.
    Our algorithm can be extended to efficient top-$K$ enumeration.


  \item By experiments on real datasets, we evaluate the efficiency and the effectiveness of our algorithm.
    We also show that there exist several models with different prediction results and fairness score
    although they have almost equal objective function values.
\end{enumerate}

\subsubsection*{Related Work}
\textit{Model enumeration} attracts increasing attention in recent years.
Enumeration algorithms for several machine learning models,
such as Lasso~\cite{hara:maehara:aaai:2017}, decision trees~\cite{Ruggieri:ICML:2017},
and rule models~\cite{hara:ishihata:AAAI:2018}, have been proposed.
In addition, a method for simultaneously learning multiple diverse classifiers
has been proposed~\cite{ross2018learning}.


\textit{Example-based explanations} are widely used for interpreting the distribution of a dataset.
Several methods for selecting representative examples from a dataset,
such as prototypes~\cite{CIS-260113} and criticisms~\cite{NIPS2016_6300}, have been proposed.
However, our method is different from theirs since
our method is based on support vectors that represent an SVM model,
and enumerates them in the descending order of the objective value.

In the context of SVM, \textit{solution path}~\cite{Hastie:2004:jmlr} is
a method for tracing changes of obtained models by varying its regularization parameter monotonically.
It is similar to our problem since it considers generation of different SVM models.
However, our problem is different from it since
our problem fixes the regularization parameter unlike a solution path algorithm varies it,
and our algorithm outputs more various models.
The \textit{uniqueness of the SVM solution} were discussed by~\cite{burges2000uniqueness}.

\section{Preliminaries}
Let $\mathbb{R}$ and and $\mathbb{N} = \set{1,2,3,\dots}$
be the sets of all real numbers and all positive integers, respectively.
For any
$n \in \mathbb{N}$, we denote by $[n] := \set{ 1, \dots, n }$.
For any indexed set $S = \set{s_1, \dots, s_n }$ and
index subset $I \subseteq [n]$,
the \name{subset of $S$ indexed by} $I$ is defined by
$S_I = \set{ s_j \mid j \in I }$.
We denote by ${\cal X}$ and ${\cal Y}$ the \textit{input} and \textit{output domains}, respectively.
In this paper, we assume the \textit{binary classification},
i.e., ${\cal X} = \mathbb{R}^{d}$ and ${\cal Y} = \set{-1,1 }$ for
some $d \in \mathbb{N}$.
A \textit{dataset} of size $n \in \mathbb{N}$ is a finite set $S = \set{(x_j, y_j)}_{j=1}^{n} \subseteq {\cal X} \times {\cal Y}$.
A
\textit{model} is any function $m : {\cal X} \to {\cal Y}$,
and a \textit{model space} is any set of models.
For other definitions, see, e.g., \cite{hastie:esl:2001}.

\subsection{Support Vector Machines (SVM)}

In the following discussion, we assume as hyperparameters
a \name{positive definite kernel function} $K: \mathbb{R}^{d} \times \mathbb{R}^{d} \to \mathbb{R}$
and a positive number $C > 0$, called a \name{regularization parameter}.
In the following, we fix $K$, $C$, and $S \subseteq {\cal X} \times {\cal Y}$, and omit them if it is clear from context.
Note that our results are independent
of
the choice of $K$ and $C$.

In this paper, we consider the dual form of SVMs~\cite{cristianini:2000:SVM}.
We assume a given dataset $S = \set{(x_j, y_j)}_{j=1}^{n}$.
For any $n$-dimensional vector $\alp = (\alpha_1, \dots, \alpha_n) \in \mathbb{R}^{n}$,
the \name{objective function} of SVMs, $f(\alp) := f(\alp \mid S, K)$, is defined by
\begin{align}
  f(\alp)
  &:= \sum_{j \in [n]} \alpha_j - \frac{1}{2}\sum_{i,j \in [n]}\alpha_i\alpha_j y_i y_j K(x_i, x_j).
  \label{eq:objective:normal}
\end{align}
The \name{feasible solution space} (or the \name{model space}) for SVMs, ${\cal F} = {\cal F}(S, C)$, is defined by the set of all Lagrange multipliers $\alp \in \mathbb{R}^{n}$
satisfying the conditions (i) and (ii) below:
\begin{equation}
  (i) \sum_{j \in [n]}\alpha_j y_j = 0
  \mbox{ and }
  (ii) \; \alpha_j \!\in\! [0,C], \forall j\in [n].
  \label{eq:feasible:solutions:normal}
\end{equation}

Now, the \name{(ordinary) SVM learning problem} is
stated
as the following maximization problem:
\begin{align}\label{max:1}
  \alp^{*} =
  \arg\max_{\alp \in {\cal F}} f(\alp).
\end{align}
Since the problem of Eq.~(\ref{max:1}) is a convex quadratic programming problem,
the solution found is global, not necessarily unique, and one of them can be efficiently computed
by various methods such as SMO~\cite{Platt:1999:FTS:299094.299105}.

By using $\alp \in {\cal F}$,
the \name{prediction model} (or the \name{SVM model}) $m : {\cal X} \to {\cal Y}$ associated to $\alp$ is
given by
\begin{align}\label{eq:model}
  m(x \mid \alp) &= m(x \mid \alp, S, K) \nonumber \\
  &:= {\rm sgn}(\sum_{j \in [n]} \alpha_j y_j K(x_j, x) + b),
\end{align}
where $x \in {\cal X}$, and a threshold $b \in \mathbb{R}$
is determined by $b = 1 - \sum_{j \in [n]} \alpha_j y_j K(x_j, x_i)$
for any $i \in [n]$
such that
$\alpha_i \in (0, C)$.
Since a model $m(\cdot \mid \alp)$ is solely determined by $\alp$,
we also call $\alp \in {\cal F}$ a \name{model} as well as $m$.

It is known that an optimal solution $\alp^{*}$ for an SVM tend to be a sparse vector.
For any $\alp \in {\cal F}$, we denote its \textit{support} and \textit{support vectors} by
$\supp(\alp) := \set{j \in [n] \mid \alpha_j \not= 0 }$ and
$V(\alp \mid S) := \set{(x_j,y_j) \in S \mid j \in \supp(\alp) }$, respectively.
From Eq.~(\ref{eq:objective:normal}), we have the next lemma, which says that the value of the objective function depends only on $\supp(\alp)$.

\begin{lemm}\label{lemm:obj:dependson:supp}
For any $\alp \in {\cal F}$ such that $\supp(\alp) \subseteq I$ for some $I \subseteq [n]$,
$f(\alp \mid S, K) = f(\alp_I \mid S_I, K)$.
\end{lemm}
\begin{prf}
  Since $\alpha_j = 0$ for any $j \in ([n] \setminus I)$, we have
  $f(\alp \mid S, K)
  = \sum_{j \in [n]} \alpha_j - \frac{1}{2}\sum_{i,j \in [n]}\alpha_i\alpha_j y_i y_j K(x_i, x_j)
  = \sum_{j \in I} \alpha_j - \frac{1}{2}\sum_{i,j \in I}\alpha_i\alpha_j y_i y_j K(x_i, x_j)
  = f(\alp_I \mid S_I, K)$.
\end{prf}


From Eq.~(\ref{eq:model}), we also see that the prediction result of SVM model $\alp$ depends only on the set of its support vectors.




\section{Problem Formulation}
Before
introducing our enumeration problem for SVMs,
we define the constrained SVM learning problem
below.
For any index subset $I \subseteq [n]$,
the \name{constrained problem associated to $I$}
is the problem~(\ref{max:1}) with
the constraint
$\supp(\alp) \subseteq I$.
Note that the problem~(\ref{max:2})
is equivalent to the problem~(\ref{max:1})
when the input is restricted to the subset $S_I \subseteq S$.

\begin{defi}
For any given index subset $I \subseteq [n]$,
the \textit{constrained SVM learning problem} with respect to $I$
is expressed as the following maximization problem:
\begin{align}\label{max:2}
  \hat\alp =
  \arg\max_{\alp \in {\cal F}(I)} f(\alpha)
\end{align}
where
${\cal F}(I) = {\cal F}(I \mid S, C)$ is the \name{constrained model space} (or the feasible solution space) consisting of all Lagrange multipliers $\alp \in \mathbb{R}^{n}$ satisfying the conditions (i) and (ii) of Eq.~(\ref{eq:feasible:solutions:normal}) and the additional condition (iii) $\supp(\alp) \subseteq I$.
\end{defi}


In the above definition,
the solution $\hat \alp$ is called a \name{support vector w.r.t.~$I$}.
We remark that the value $f(\alp)$ does not depend on the choice of $I$
since $f(\alp) = f(\alp \mid S, K) = f(\alp_I \mid S_I, K)$
by condition (iii) and Lemma~\ref{lemm:obj:dependson:supp}.

Then,
we denote the set of \name{globally optimal solutions} by
\begin{math}
  \sol(I) = \sol(I \mid S,C,K) := \set{ \alp \in {\cal F}(I) \mid f(\alp) = {\hat f} },
\end{math}
where ${\hat f} = \max_{\alp \in {\cal F}(I)} f(\alpha)$ is the optimum value for the objectives.

The following property plays a key role in the analysis of our algorithm proposed later.

\begin{prop}[key property of solutions]\label{prop:1}
  Let $I \subseteq [n]$ be any index subset and $\halp \in \sol(I)$ be any solution w.r.t.~$I$.
  For any $J \subseteq [n]$, $\supp(\halp) \subseteq J \subseteq I \implies \halp \in \sol(J)$.
\end{prop}
\begin{prf}
  By assumption, we have (a) $\supp(\halp) \subseteq J$ implies $\halp \in {\cal F}(J)$, and (b) $J\subseteq I$ implies ${\cal F}(J) \subseteq {\cal F}(I)$.  From (a) and (b), we have that if $\halp \in \sol(I)$ is optimal in ${\cal F}(I)$, it is also optimal in ${\cal F}(J)$. Thus, $\halp \in \sol(J)$ is proved.
\end{prf}

An \name{algorithm for the constrained SVM problem} is
any deterministic algorithm $\svm: 2^{[n]} \to {\cal F}$
that given $I\subseteq [n]$ as well as $S, K$, and $C$, computes a solution $\alp = \svm(I) \in \sol(I)$ for the SVM problem.
From Proposition~\ref{prop:1}, we make the following assumption on $\svm$ throughout this paper.

\begin{assu}
  \label{assu:svmalgo}
  For any $I,J \subseteq [n]$, $\svm$ satisfies that $\supp(\svm(I))$ $\subseteq J \subseteq I$ implies $\svm(I) = \svm(J)$.
\end{assu}

For justification of Assumption~\ref{assu:svmalgo}, if the objective function~$f$ is strictly convex, the set $\sol(I)$ has the unique solution~\cite{burges2000uniqueness}, and thus, the assumption holds.
If $f$ is not strictly convex,
it is only known that $\sol(I)$ is itself a convex set~\cite{burges2000uniqueness}.
It remains open if a sort of greedy variable selection strategies in,
e.g., the SMO~\cite{Platt:1999:FTS:299094.299105} or the chunking~\cite{Vapnik:1998} algorithms
is sufficient to ensure Assumption~\ref{assu:svmalgo}.

Under the above assumption, the solution set for our enumeration problem on input $(S, C, K)$ is the collection
${\cal M}_{\rm all}
= \set{\alp \in {\cal F} \mid \exists  I \subseteq [n]: \alp = \svm(I) }$
of the distinct SVM models computed by $\svm$ for all possible index subsets of $[n]$. We observe that the corresponding set
${\cal SV_{\rm all}} = \set{\supp(\alp) \mid \alp \in {\cal M}_{\rm all}}
\subseteq 2^{[n]}$
of the supports is isomorphic to the quotient set $\Pi$ w.r.t.~the equivalence relation $\equiv_{\svm}$
defined by
\begin{math}
I \equiv_{\svm} J \iff \svm(I) = \svm(J),
\end{math}
where each representative
$[J] \in \Pi$
is written as
$[J] = \set{ I \in 2^{[n]} \mid I \equiv_{\svm} J }$.

Our goal
is to enumerate all models of ${\cal M}_{\rm all}$ that have distinct support vectors in the descending order of their objective function values.
Now, we state our problem as follows.

\begin{prob}[Enumeration problem for SVMs]\label{prob:enum}
  Given any dataset $S=\set{(x_j, y_j)}_{j=1}^{n} \subseteq {\cal X}\times{\cal Y}$, parameter $C>0$, and kernel function $K$, the task is to enumerate all distinct models $\alp$ in ${\cal M}_{\rm all}$
  in the descending order of their objective function values $f(\alp)$ without duplicates.
\end{prob}


Note that we fix the regularization parameter $C$
unlike the solution path for SVMs~\cite{Hastie:2004:jmlr}.

To solve Problem~1, a straightforward, but infeasible method is
to simply collect $\svm(I)$
over all exponentially many subsets $I$ in $2^{[n]}$.
This has redundancy w.r.t.~$\equiv_{\svm}$ since some pair of subsets $I$ and $J$ may yield the same solution if they are equivalent.
Hence, we seek for a more efficient method utilizing the sparseness of the SVM models in ${\cal M}_{\rm all}$.


\newcommand{\algo}[1]{\textsc{#1}}

\section{Algorithm}

In this section, we propose an efficient algorithm
\algo{EnumSV}
for solving Problem~\ref{prob:enum}.
\algo{EnumSV}
is based on Lawler's framework~\cite{lawler:1972::kbest} for top-K enumeration
following the approach by Hara and Maehara to Lasso~\cite{hara:maehara:aaai:2017}.

\begin{algorithm}[t]
  \caption{An enumeration algorithm \algo{EnumSV} for SVMs that,
    given
    a dataset $S=\set{(x_j, y_j)}_{j=1}^{n}$, $C>0$, and a kernel function $K$,
           returns the set of all models in ${\cal M}_{\rm all} = \set{\svm(I \mid S,C,K) \mid I \subseteq [n]}$
           in descending order of their objective function values without duplicates.
           }
  \label{algo:1}
  \begin{algorithmic}[1]
    \STATE Heap $H \leftarrow \emptyset$;
    \STATE $\halp \leftarrow \svm([n] \mid S, C, K)$;
    \STATE Insert $\tau_1 = (\halp, [n], \emptyset)$ into the heap $H$;
    \STATE ${\cal M} \leftarrow \emptyset$;
    \WHILE{$H \not= \emptyset$}
    \STATE Extract $\tau = (\alp, I, B)$ from the heap $H$;
    \label{line:solcand}
    \label{line:xxx}
      \STATE \textbf{if} $\alp \not\in {\cal M}$ \textbf{then}
         ${\cal M} \leftarrow {\cal M} \cup \set{ \alp }$; \label{line:solin}
      \FOR{$j \in \supp(\alp)\setminus B$} \label{line:for}
        \STATE $I' \leftarrow I \setminus \set{ j }$; \label{line:branch}
        \STATE $\alp' \leftarrow \svm(I' \mid S, C, K)$; \label{line:comp}
        \STATE Insert $\tau' = (\alp', I', B)$ into the heap $H$; \label{line:insert}
        \STATE $B \leftarrow B \cup \set{ j }$; \label{line:fset}
      \ENDFOR
    \ENDWHILE
    \STATE {\bfseries Return} ${\cal M}$; \label{line:ret}
  \end{algorithmic}
\end{algorithm}

\subsection{The outline of our algorithm}
In Algorithm~\ref{algo:1}, we show the outline of our algorithm \algo{EnumSV}.
It maintains as a data structure $H$, which is
a \name{priority queue} (or a \textit{heap})~\cite{Cormen:2009:IAT:1614191},
to store triples
$\tau = (\alp, I, B)$
consisting of
\vspace{-2.5mm}
\begin{itemize}
  \itemsep=0pt
\item a discovered solution $\alp \in {\cal F}$ (a Lagrange multiplier),
\item an index set $I \in 2^{[n]}$ associated to $\alp$ by $\alp = \svm(I)$, and
\item a forbidden set $B \in 2^{[n]}$
  to avoid searching redundant children of $I$.
\end{itemize}

Triples are ordered in the descending order of their objective values $f(\alp)$ as keys.
For the heap $H$, we can \name{insert} to $H$ any triple and \name{extract} (or \name{deletemax}) from $H$ the triple $\tau$ with the maximum key each in $O(\log |H|)$ time~\cite{Cormen:2009:IAT:1614191}.

In Lawler's framework, we can compute an optimal solution for each subproblems avoiding subproblems that yields redundant solutions.

\textbf{Base Case}: Initially, \algo{EnumSV} starts by inserting  the first triple
$\tau_1 = (\svm([n]), [n], \emptyset)$
at Line~3, where $\tau_1$ corresponds to the solution for the ordinary SVM problem.

\textbf{Inductive Case}: While $H$ is not empty, \algo{EnumSV} then repeats the following steps in the while-loop:
\vspace{-2.5mm}
\begin{description}
  \itemsep=0pt
  \item[Step~1] Extract a triple $\tau = (\alp, I, F)$ from the heap $H$ at Line~6, where $\alp$ is called a \textit{candidate}.
        Insert vector $\alp$ to ${\cal M}$ as a solution at Line~7 if it has not been founded yet.
  \item[Step~2] Repeat the following steps for any
        $j \in \supp(\alp) \setminus B$:
  \begin{enumerate}
  \itemsep=0pt
    \item Branch the search spaces as $I' = I \setminus \set{j}$ at Line~9.
    \item Compute $\alp' = \svm(I')$ at Line~10
          and insert the triple $\tau' = (\alp', I', F)$, called a \name{child} of $\tau$, into the heap $H$ at Line~11.
    \item Insert $j$ into $B'$ to avoid inserting the same index subset into the heap $H$ twice at Line~12.
  \end{enumerate}
  \item[Step~3] Back to step 1. if the heap $H$ is not empty.
\end{description}

The most important step of \algo{EnumSV} in Algorithm~\ref{algo:1} is
Step~2
above.
Based on Proposition~\ref{prop:1}, it branches a search
on each index $j \in \supp(\alp) \setminus B$.
We can avoid redundant computations that yield the same solution that had already been output before.
To avoid enumerating the same index subset $I$ multiple times,
we add the used index $j$ into $B$.

\subsection{The correctness}

In this subsection, we show the correctness of \algo{EnumSV} in Algorithm~\ref{algo:1} on input $(S, C, K)$ through properties~1 and~2 below.
For every $k\ge 1$, $\alp^{(k)}$ denotes the $k$-th solution in ${\cal M}$ by \algo{EnumSV}.
We first show a main technical lemma.

\begin{lemm}\label{lemm:1}
  For any feasible solution $\alp \in {\cal F}$, there exists some
  $\tau^{(k)} = (\alp^{(k)}, I^{(k)}, B^{(k)})$
  extracted from $H$
  such that
  (i) $\supp(\alp^{(k)}) \subseteq \supp(\alp)$,
  (ii) $\supp(\alp) \subseteq I^{(k)}$, and
  (iii) $f(\alp^{(k)}) \geq f(\alp)$, where $k\ge 1$.
\end{lemm}

\begin{prf}
  Let $\alp \in {\cal F}$.
    Starting from the initial triple $\tau^{(1)}$, we will go down the search space by visiting triples $\tau^{(k)}$
  from a parent to its child for $k = 1,2,\dots$, while
  maintaining the invariant (ii).
  Base case: For $k = 1$, the first triple $\tau^{(1)}$ clearly satisfies
  the invariant $\supp(\alp) \subseteq I^{(k)}$.
  From Lemma~\ref{lemm:obj:dependson:supp}, if $\tau^{(1)}$ satisfies condition (i), the claim immediately follows.

  Induction case:
  Let $k > 1$. Suppose inductively that
  $\tau^{(k)}$ satisfies (ii) $\supp(\alp) \subseteq I^{(k)}$.
  Then, there are two cases (1) and (2) below on the inclusion $\supp(\alp^{(k)}) \subseteq \supp(\alp)$:

  Case (1): $\supp(\alp^{(k)}) \subseteq \supp(\alp)$ holds.
  By induction hypothesis, we have $\supp(\alp^{(k)}) \subseteq \supp(\alp) \subseteq I^{(k)}$, and thus, $\alp \in {\cal F}(S \mid I^{(k)})$.
  Since $\alp^{(k)}$ is an optimal solution within ${\cal F}(S \mid I^{(k)})$, it follows that $f(\alp^{(k)}) \geq f(\alp)$.
  Case (2): $\supp(\alp^{(k)}) \not\subseteq \supp(\alp)$ holds.
  For any $j \in \supp(\alp^{(k)})\setminus \supp(\alp)$,
  \algo{EnumSV} inserts into the heap the triple $\tau' = (\alp', I', B')$ at Line~11, and it will be eventually extracted as the $m$-th triple $\tau' = \tau^{(m)} = (\alp^{(m)}, I^{(m)}, F^{(m)})$ at some $m > k$.
  By induction hypothesis, $\supp(\alp) \subseteq I^{(k)}$ and $j \not\in\supp(\alp)$ hold, and thus, we have an invariant $\supp(\alp) \subseteq I^{(k)}\setminus\set{j} = I^{(m)}$ for the child iteration with $\tau^{(m)}$.
  By the above arguments, at every time following a path to a child in Case (2), the size of the difference $\Delta^{(m)} = |I^{(m)} \setminus \supp(\alp)|$ decrements at least by one. Since $\Delta^{(m)} \ge 0$, this process must eventually halt at Case (1). This completes the proof.
\end{prf}

From Lemma~\ref{lemm:1}, we can show the next lemma, saying that \algo{EnumSV} eventually outputs any solution.

\begin{lemm}[Property~1]
  \label{lemm:2}
  In \algo{EnumSV}, for any subset $I \subseteq [n]$, there exists some $k\ge 1$  such that $\alp^{(k)} = \svm(I)$.
\end{lemm}
\begin{prf}
  For $\alp' = \svm(I)$, it follows from Lemma~\ref{lemm:1} that there exists $k \in \mathbb{N}$ such that
  $\supp(\alp^{(k)}) \subseteq \supp(\alp')$ and $f(\alp^{(k)}) \geq f(\alp')$.
  Since $\alp^{(k)} \in {\cal F}(I)$, we have $\alp^{(k)} = \svm(I)$.
\end{prf}

Also from Lemma~\ref{lemm:1}, we have the next lemma
for the top-$K$ computation, which says that \algo{EnumSV} lists solutions $\alp$ exactly from larger to smaller values of $f(\alp)$.

\begin{lemm}[Property~2]
  \label{lemm:3}
  \algo{EnumSV}
  enumerates solutions $\alp$
  in the descending order of their objective function
  $f(\alp)$,
  i.e., $f(\alp^{(1)}) \ge \dots \ge f(\alp^{(k)}) \ge \dots\;(k\ge 1)$.
\end{lemm}

\begin{prf}
  We show $f(\alp^{(k)}) \geq f(\alp^{(m)})$ for any $m > k$ as follows.
  Suppose that $\alp^{(k)}$ is extracted by \name{deletemax} from the heap at step $k$.
  If $\alp^{(m)}$ is in the heap, then $f(\alp^{(k)}) \geq f(\alp^{(m)})$ immediately holds.
  Otherwise, there exists the triple $(\alp^{(\ell)}, I^{(\ell)}, B^{(\ell)})$
  where $\ell < m$ in the heap such that $I^{(m)} \subseteq I^{(\ell)}$.
  Since $\alp^{(m)} \in {\cal F}(S \mid I^{(\ell)})$, $f(\alp^{(\ell)}) \geq f(\alp^{(m)})$ holds.
  From the definition of the heap, we have $f(\alp^{(k)}) \geq f(\alp^{(\ell)}) \geq f(\alp^{(m)})$.
\end{prf}

By combining Lemma~\ref{lemm:2} and Lemma~\ref{lemm:3}, we show the main result of this paper.

\begin{theo}\label{theo:comp}
  \algo{EnumSV} in Algorithm~\ref{algo:1} solves Problem~\ref{prob:enum}.
\end{theo}

\begin{prf}
  From Lemmas~\ref{lemm:2} and \ref{lemm:3},
  \algo{EnumSV} returns a collection of models ${\cal M} = \set{\alp^{(1)}, \alp^{(2)}, \dots}$
  that satisfy Properties 1 and 2.
  Thus, \algo{EnumSV} solves Problem~\ref{prob:enum}.
\end{prf}

\subsection{Top-$K$ Enumeration}

We can
modify Algorithm~\ref{algo:1} to find the top-$K$ models
${\cal M}_{K} = \set{\alp^{(1)}, \dots, \alp^{(K)}} \subseteq {\cal M}_{\rm all}$
for a given positive integer $K \in \mathbb{N}$ as follows.
We simulate Algorithm~\ref{algo:1},
perform the enumeration of models in the descending order of their objectives, and then stop Algorithm~\ref{algo:1} when $|{\cal M}| = K$ eventually holds.
From Lemma~\ref{lemm:3}, we see that ${\cal M}_{K} \subseteq {\cal M}_{\rm all}$ and ${\cal M}_{K}$ contains the top-$K$ models.

\paragraph{Complexity. }
For enumeration algorithms,
it is the custom to analyze their time complexity
in terms of the number of solutions, or in \textit{output-sensitive} manner~\cite{avis1996reverse}.
However, it is difficult
because more than one equivalent candidates $I \not= J$ can result the model $\svm(I) = \svm(J)$.
Instead, we estimate its time complexity in terms of a candidate solution extracted in Line~6.
The time complexity of Algorithm~\ref{algo:1} for obtaining a candidate solution of $k$-th solution $\alp^{(k)}$
is ${\cal O}({\cal T}_\textrm{SVM} \cdot |\supp(\alp^{(k-1)})|)$, where ${\cal T}_\textrm{SVM}$ is the complexity of solving an SVM problem.

\section{Experiments}
In this section, we evaluate our algorithm by experiments on real datasets.
All codes were implemented in Python~3.6 with scikit-learn.
We used linear kernel $K(x_i, x_j)=x_{i}^{\rm T} x_j$ as the kernel function in all experiments.
All experiments were conducted on 64-bit Ubuntu 18.04.1 LTS with Intel Xeon E5-1620 v4 3.50GHz CPU and 62.8GiB Memory.

\begin{figure}[t]
  \centering
  \subfigure[Objective function value ratio $f(\alp^{(k)})/f(\alp^{(1)})$.]{
      \includegraphics[width=\columnwidth]{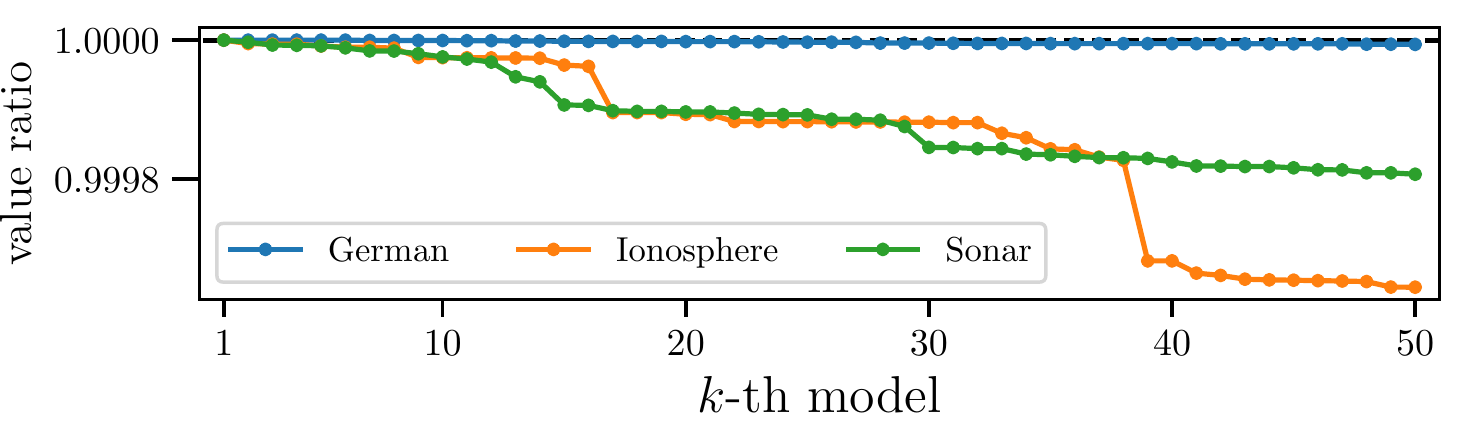}
  }
  \subfigure[Test loss ratio of $\alp^{(k)}$ on the test dataset.]{
      \includegraphics[width=\columnwidth]{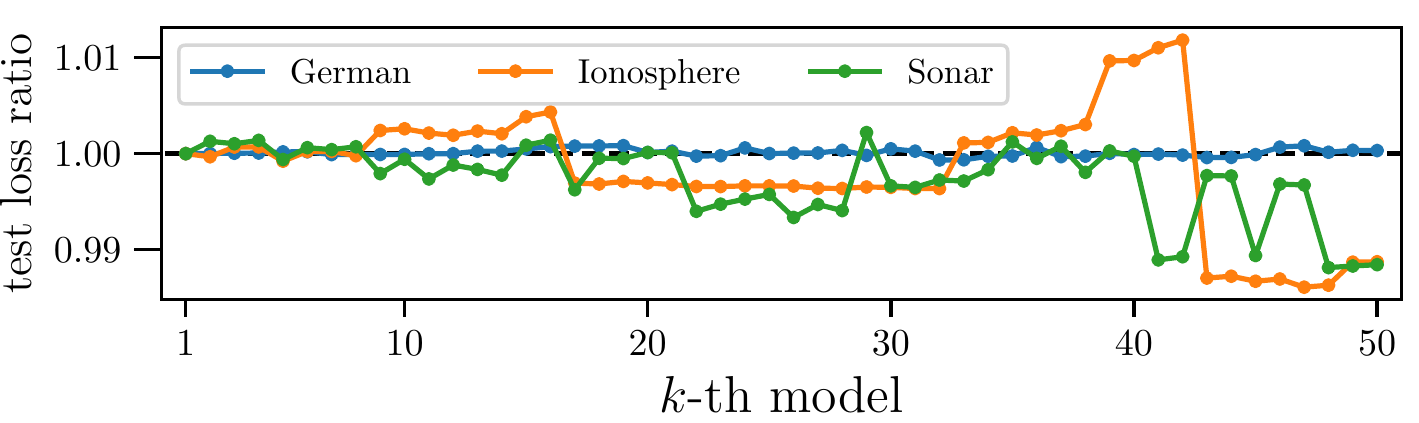}
  }
  \caption{Changes of objective function value and test loss ratio of the top-50 enumerated models on UCI datasets.}
  \label{fig:1}
\end{figure}

\begin{figure}[t]
  \centering
  \centerline{\includegraphics[width=\columnwidth]{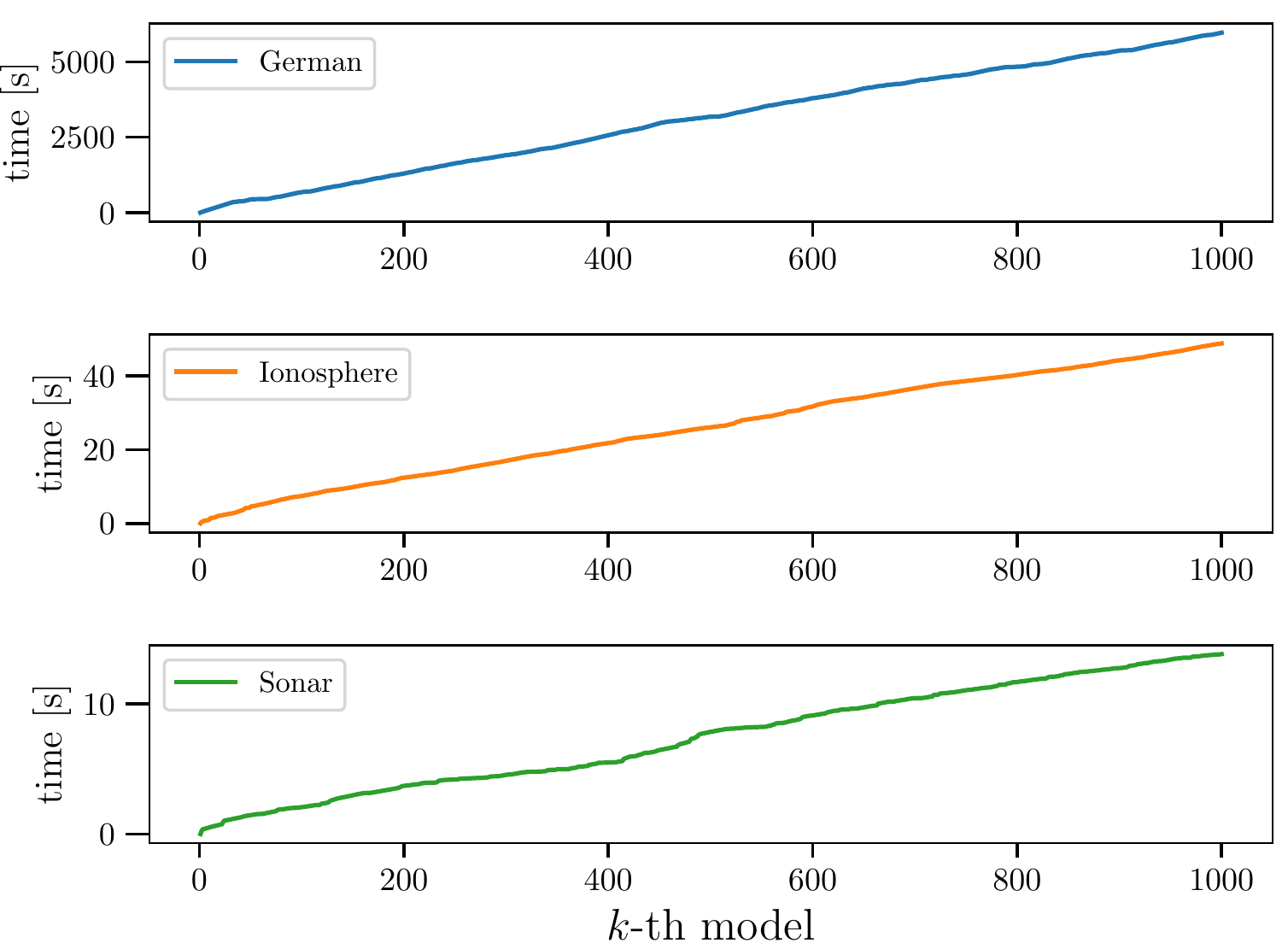}}
  \caption{Running times for UCI datasets.}
  \label{fig:2}
\end{figure}

\subsection{UCI Datasets}

We first evaluated \algo{EnumSV} on three real datasets,
German ($n=1000, d=20$), Ionosphere ($n=351, d=34$), and Sonar ($n=208, d=60$) from UCI ML repository~\cite{Dua:2017}.
Their task is a binary classification.
We randomly split each dataset into train ($70\%$) and test ($30\%$) samples,
and evaluated the test loss by the hinge loss $l(y, {\hat y}) = \max (0, 1-y \cdot {\hat y})$.
For each dataset, the hyperparameter $C$ was selected by $5$-fold cross validation among $10^{-2}, 10^{-1}, \dots, 10^{3}$
before enumeration.

We applied \algo{EnumSV} to these datasets, and enumerated top-$50$ models.
Figure~\ref{fig:1} presents the values of the ratio $f(\alp^{(k)})/f(\alp^{(1)})$ of the objective function value and the ratio of the test loss
of the $k$-th enumerated model $\alp^{(k)}$ to those of the best model $\alp^{(1)}$.
Figure~\ref{fig:1}~(a) shows that the values of the objective function decreases as the rank $k$ increases
as expected from Theorem~\ref{theo:comp}.
For German dataset, the objective function values of top-$50$ were almost same within deviation of $6.0 \times 10^{-4}$.
It indicates that there are multiple models achieving the almost identical objective value.
Figure~\ref{fig:1}~(b) shows that some enumerated model, such as $\alp^{(32)}$ for German,
$\alp^{(43)}$ for Ionosphere, and $\alp^{(41)}$ for Sonar, had smaller test loss compared with the optimal model $\alp^{(1)}$.
It means that an optimal model is not always the best model, and
we obtained a better model with a lower test loss than an optimal model by enumerating models.

Figure~\ref{fig:2} presents the total time of enumerating top-$1000$ models.
The total time seems almost linear in rank $k$.
Consequently, we conclude that \algo{EnumSV} has small latency for outputting solutions independent of their ranks, and thus, is scalable in the number $K$ of enumerated models.

\subsection{Injected COMPAS Dataset}
\begin{figure}[t]
  \centering
  \subfigure[Demographic parity $\delta(\alp^{(k)} \mid z)$ on the test dataset $S_\textrm{test}$.]{
      \includegraphics[width=\columnwidth]{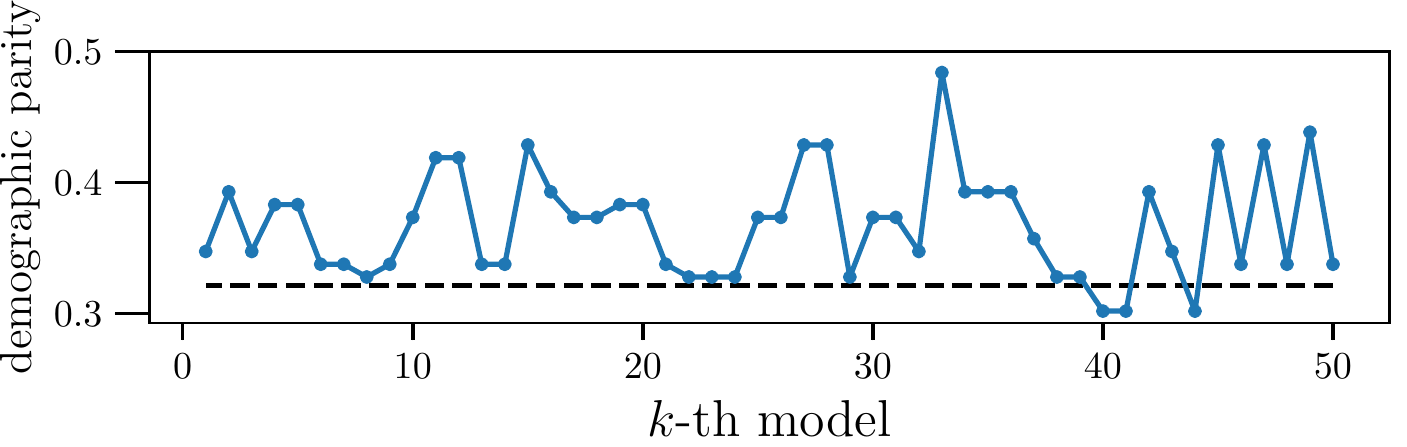}
  }
  \subfigure[Misclassification ratio of $\alp^{(k)}$ on $S_\textrm{test}$.]{
      \includegraphics[width=\columnwidth]{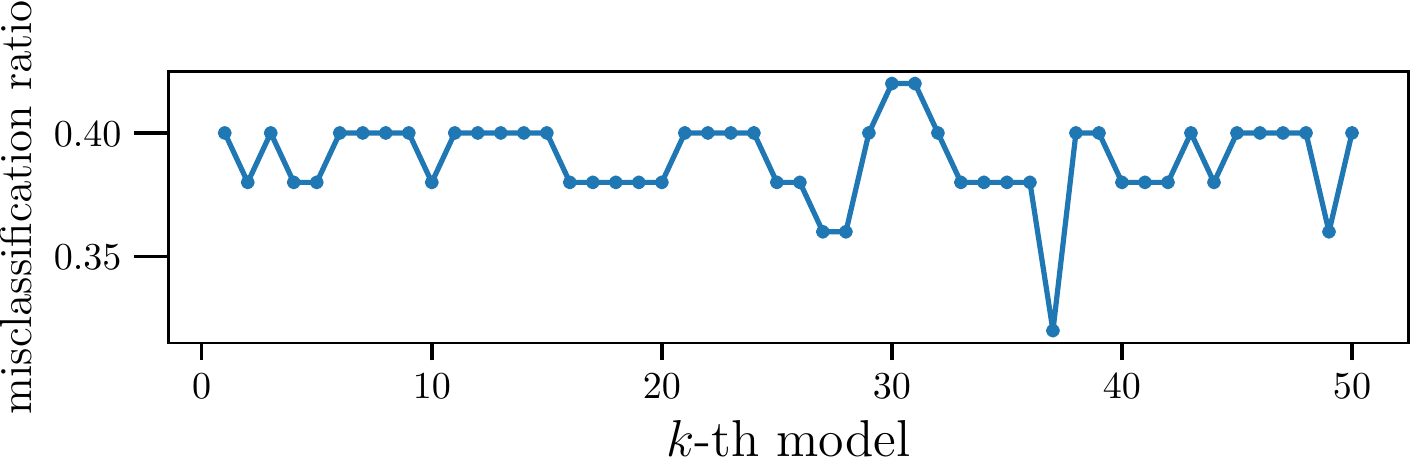}
  }
  \caption{Changes of properties of top-50 enumerated models on Injected COMPAS dataset.}
  \label{fig:3}
\end{figure}

Next, we demonstrate an application of \algo{EnumSV}
to a fair classification scenario under \textit{false data injection attacks}.
To evaluate the fairness of the model $m$ for the sensitive attribute $z \in \set{-1, 1}$,
we used \textit{demographic parity (DP)}~\cite{5360534} defined by
\begin{align*}
  \delta(m \!\mid\! z) := | P(m(x) \!=\! 1 \!\mid\! z \!=\! 1) - P(m(x) \!=\! 1 \!\mid\! z \!=\! -1) |,
\end{align*}
where $P$ is a probability on the joint distribution over $(z, m(x))$.
We note that the larger the DP, the larger the discrimination of prediction.

We used COMPAS dataset ($n=6172, d=12$) related to recidivism risk prediction distributed at \cite{adebayo:fairml}.
The task is to predict whether individual people recidivate within two years from their criminal history.
We used the attribute "African\_American" as a sensitive attribute $z$.
We assume a scenario of \textit{false data injection}~\cite{5718158} that is a special kind of attacks to learning algorithms,
which increases the DP of the learned model for the sensitive attribute $z$
by flipping output labels $y$ of a small subset of a training dataset.
To reproduce this scenario, we generated injected subsets of the COMPAS by the following steps:
\begin{enumerate}
  \itemsep=0pt
  \item Create a training dataset $S$ by randomly sampling a subset of the COMPAS with $100$ examples.
  \item Randomly choose a subset of $S$ such that $y \not= z$ with $10$ examples, and replace these outputs $y$ by $-y$.
  \item Create a test dataset $S_\textrm{test}$ by randomly sampling from the COMPAS with $50$ examples.
\end{enumerate}
By our preliminary experiments, we confirm that the above procedure increases the DP of SVM models on $S_\textrm{test}$.

We applied \algo{EnumSV} to the above injected COMPAS dataset,
and measured {objective function values}, \textit{demographic parity (DP)}, and \textit{misclassification ratio}
of the top-50 enumerated models.
We observed that all the enumerated top-$K$ models had the same objective value.
However, these prediction results were mutually different.

Figure~\ref{fig:2} (a) presents the value of the DP of the enumerated models,
where the dashed line indicates the reference DP value $\theta_{*} = 0.321$ of
the model learned by the non-injected subset of the input $S$.
Figure~\ref{fig:2} (b) presents the misclassification ratio of the enumerated models $\alp^{(k)}$ on $S_\textrm{test}$.
From the figures, we observed that \algo{EnumSV} found
the three fair models $\alp^{(40)}$, $\alp^{(41)}$, and $\alp^{(44)}$
achieving lower DP than~$\theta_{*}$ and lower misclassification ratio than $\alp^{(1)}$.
Consequently, \algo{EnumSV} successfully obtained several fair models against false data injection by enumerating models.


\section{Conclusion}
In this paper,
we proposed an efficient algorithm to enumerate top-$K$ SVM models with distinct support vectors
in descending order of these objective function values.
By experiments on real datasets, we demonstrated that
our
framework provides
better models than one single optimal solution, and
fair models against false data injection, which increases the unfairness of an optimal model.
As future work,
we will try to make theoretical or empirical justification of Assumption~\ref{assu:svmalgo} for a particular class of SVM learning algorithms such as chunking~\cite{Vapnik:1998} and SMO~\cite{Platt:1999:FTS:299094.299105}.
It is also interesting future work
to extend our algorithm to enumerate models taking their diversity into account
so as to interactively help users to understand a dataset.


\section*{Acknowledgements}
This work was partially supported by JSPS KAKENHI(S) 15H05711 and JSPS KAKENHI(A) 16H01743.

\bibliography{kanamori}
\bibliographystyle{icml2019}

\end{document}